\documentclass[sn-mathphys-num]{sn-jnl}% Math and Physical Sciences Numbered Reference Style 
\usepackage{graphicx}%
\usepackage{multirow}%
\usepackage{amsmath,amssymb,amsfonts}%
\usepackage{amsthm}%
\usepackage{mathrsfs}%
\usepackage[title]{appendix}%
\usepackage{xcolor}%
\usepackage{textcomp}%
\usepackage{manyfoot}%
\usepackage{booktabs}%
\usepackage{algorithm}%
\usepackage{algorithmicx}%
\usepackage{algpseudocode}%
\usepackage{listings}%

\theoremstyle{thmstyleone}%
\theoremstyle{thmstyletwo}%

\theoremstyle{thmstylethree}%

\begin{document}

\title[MEDPNet: Achieving High-Precision Adaptive Registration for Complex Die Castings]
{MEDPNet: Achieving High-Precision Adaptive Registration for Complex Die Castings}

\author[1]{\fnm{Yu} \sur{Du}}

\author[1]{\fnm{Yu} \sur{Song}}

\author[2]{\fnm{Ce} \sur{Guo}}

\author*[1]{\fnm{Xiaojing}\sur{Tian}}\email{tzy@djtu.edu.cn}

\author[2]{\fnm{Dong} \sur{Liu}}

\author[2]{\fnm{Ming} \sur{Cong}}

\affil*[1]{\orgdiv{School of Mechanical Engineering}, \orgname{Dalian Jiaotong University}, \orgaddress{\street{794 Huanghe Road}, \city{Dalian}, \postcode{116028}, \state{Liaoning}, \country{China}}}

\affil[2]{\orgdiv{School of Mechanical Engineering}, \orgname{Dalian University of Technology}, \orgaddress{\street{2 Linggong Road}, \city{Dalian}, \postcode{116024}, \state{Liaoning}, \country{China}}}

\abstract{Due to their complex spatial structure and diverse geometric features, achieving high-precision and robust point cloud registration for complex Die Castings has been a significant challenge in the die-casting industry. Existing point cloud registration methods primarily optimize network models using well-established high-quality datasets, often neglecting practical application in real scenarios. To address this gap, this paper proposes a high-precision adaptive registration method called Multiscale Efficient Deep Closest Point (MEDPNet) and introduces a die-casting point cloud dataset, DieCastCloud, specifically designed to tackle the challenges of point cloud registration in the die-casting industry.
The MEDPNet method performs coarse die-casting point cloud data registration using the Efficient-DCP method, followed by precision registration using the Multiscale feature fusion dual-channel registration (MDR) method. We enhance the modeling capability and computational efficiency of the model by replacing the attention mechanism of the Transformer in DCP with Efficient Attention and implementing a collaborative scale mechanism through the combination of serial and parallel blocks. Additionally, we propose the MDR method, which utilizes multilayer perceptrons (MLP), Normal Distributions Transform (NDT), and Iterative Closest Point (ICP) to achieve learnable adaptive fusion, enabling high-precision, scalable, and noise-resistant global point cloud registration.
Our proposed method demonstrates excellent performance compared to state-of-the-art geometric and learning-based registration methods when applied to complex die-casting point cloud data.}


\maketitle

\section{Introduction}\label{sec1}

Complex Die Castings are critical components in industries such as manufacturing, transportation, and defense, characterized by intricate structures and diverse forms. High-quality three-dimensional reconstruction of their overall surfaces through point cloud registration plays a vital role in enhancing product molding quality and ensuring safety in subsequent use. Recent work has made substantial progress in fully automatic, 3D feature-based point cloud registration. At first glance, benchmarks like 3DMatch \cite{3dmatch} appear to be saturated, with multiple state-of-the-art (SoTA) methods reaching nearly 95\% feature matching recall and successfully registering over 80\% of all scan pairs \cite{predator}.
However, due to the complexity of the spatial structure of Die Castings and their susceptibility to complex background interferences such as casting reflections, oil contamination, machining marks, etc., there is currently no effective method to achieve high-precision point cloud registration of die-casting data. We believe that a high-precision adaptive method is the key to addressing this issue.

\begin{figure}[t]
    \centering
    \includegraphics[width=0.8\linewidth]{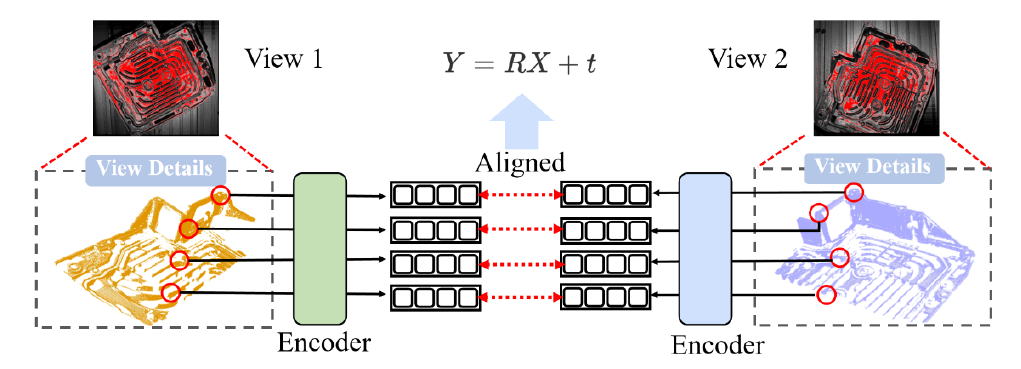}
\caption{\textbf{Illustrates the point cloud registration process for a die-cast part}. **"View 1" and "View 2" correspond to point clouds X and Y under different views.** Through an encoder, the point cloud data of the die-cast part are converted into data in feature space. Then, from the common parts of the unaligned point cloud pairs, the optimal transformation parameters {R, t} are calculated to obtain the best transformation matrix \(T^*\), where R represents the rotation angles, t represents the translation components, and \(T^*\) is the best transformation matrix.}
%\vspace{-.2cm}
\label{fig:1}
\end{figure}

Currently, representative point cloud registration methods can be broadly categorized into two main types: those based on geometric properties\cite{icp}\cite{ndt}\cite{anisotropic}\cite{CPD}] and those based on deep learning\cite{predator}\cite{hu2023bag}\cite{prnet}\cite{wu2022sample}\cite{dcp}. The method for point cloud registration, as shown in Fig \ref{fig:1}, aims to calculate the optimal transformation parameters (R, T) (three rotation angles in R and three translation components in T) from the common parts of the data known as correspondences\cite{Novel1}. In recent years, with the rapid development of deep learning\cite{hu2023robust}\cite{git}, it has found widespread application in point cloud registration tasks. Deep learning-based registration algorithms, including DCP\cite{dcp}, PointNetLK\cite{pointnetlk}, GeoTransformer\cite{geo}, etc., have significantly improved the speed and performance of point cloud registration tasks. However, these methods often require more computational resources, and their performance is frequently constrained by the quality of the dataset, often leading to suboptimal results in practical applications and difficulty in achieving stable high-precision point cloud registration effects.

Representative point cloud registration algorithms based on geometric properties include Iterative Closest Point (ICP)\cite{icp}  and Normal Distributions Transform (NDT)\cite{ndt}. 
Such methods have low hardware requirements, are easy to implement, exhibit strong interpretability, and do not involve time-consuming training processes. However, they face challenges such as sensitivity to local minima or poor generalization, reliance on manually crafted features to distinguish corresponding relationships, and significant impact from the designer's experience and parameter tuning capabilities. Additionally, these methods often consume considerable time, posing potential bottlenecks in real-time applications.

In addressing this challenge, given the intricate nature of die-cast components, we have devised a highly efficient adaptive registration method and created the DieCastCloud point cloud dataset to validate the efficacy of the approach.DCP method performs well in point cloud registration tasks, but it still has certain stability issues when faced with complex die-cast point cloud data with diverse surface feature variations. 
To address this problem, we introduce Efficient Attention\cite{efficient} to replace the Transformer Attention\cite{Transformer} in DCP. By combining serial and parallel blocks, Efficient Attention efficiently captures global feature information and improves computational efficiency. Efficient Attention differs from traditional self-attention mechanisms in terms of implementation. Traditional self-attention mechanisms\cite{weng2023cross}\cite{li2022enhancing}\cite{qiao2022novel} generate an attention map for each position to aggregate input values and produce outputs. In contrast, Efficient Attention does not generate separate attention maps for each position. Instead, it interprets the keywords of attention as global attention maps, with each global attention map corresponding to a semantic aspect of the entire input. Efficient Attention uses these global attention maps to aggregate values and generate a global context vector. 
Then, each position uses a set of coefficients to weight the global context vector and adjust its own representation. This approach gives Efficient Attention advantages in terms of memory and computational efficiency, as it does not require calculating similarities between each pair of positions, thereby reducing computational and storage complexities.
Although the improved DCP network provides initial registration for casting point clouds, it has certain limitations in terms of accuracy and stability, restricting its practical feasibility in industrial applications. 

To address this issue, we propose a multi-scale adaptive fine registration method. Building upon a favorable initial pose obtained from the DCP network, we further enhance the precise registration of point cloud data through the fusion of multi-scale feature information. In order to ensure the robustness and verifiability of fine registration, we improve and integrate ICP and NDT. Specifically, we initially perform multi-scale feature extraction on the die-casting point cloud data to avoid the impact of feature loss or noise interference on point cloud registration. Subsequently, we use a dual-channel approach to obtain transformation matrices from NDT and ICP that have undergone multi-scale feature fusion. We then apply nonlinear weighting to the obtained transformation matrices, endowing them with good adaptability and stable registration accuracy.

\begin{itemize}
  \item We replaced the Transformer's Attention in DCP with Efficient Attention and implemented a collaborative scale mechanism through a combination of serial and parallel blocks to improve both the modeling capability and computational efficiency of the model.
  \item We propose a Multiscale feature fusion dual-channel precision registration(MDR) method and supported our experimental details through ablation experiments under various scenarios.
  \item We established the point cloud dataset DieCastCloud to address the challenge of scarce high-quality point cloud data in the die casting industry.
\end{itemize}

% 图1

\section{Related Work}\label{sec2}

\textbf{DCP (Deep Closest Point):  }DCP\cite{dcp} is a representative learning-based method for point cloud registration\cite{pointnetlk}\cite{geo}\cite{deepgmr}. The primary objective of the DCP method is to address issues encountered by traditional methods such as 4PCS(4-points congruent sets for robust pairwise surface registration)\cite{4pcs} and CPD(Coherent point drift)\cite{CPD}, which are prone to noise interference and susceptible to getting trapped in local optima. DCP utilizes deep neural networks\cite{deep_neural_networks} to learn representations of point clouds and employs these learned representations for point cloud registration. Specifically, the DCP\cite{dcp} method comprises two sub-networks: a feature extraction network and a transformation prediction network. The feature extraction network is responsible for extracting local and global geometric features\cite{fu2022bag} from the input point clouds, while the transformation prediction network predicts the rigid transformation required to align two point clouds.

To tackle the matching problem in point cloud registration\cite{3dmatch}\cite{GMM}\cite{deepgmr}\cite{4pcs}\cite{ef_icp}, DCP\cite{dcp} adopts the approach of Pointer Networks\cite{pointernetworks}. Pointer Networks use attention mechanisms to select positions in the input sequence, addressing the challenge of predicting discrete labels. By predicting a position distribution at each output step, Pointer Networks can be considered as "soft pointers" for selecting matching positions. The entire network is differentiable, allowing for end-to-end training\cite{end2end}. HSGM \cite{hsgm}\cite{hsgnet} proposes a hierarchical similarity graph module to relieve the conflict of backbone networks and mine the discriminative features. Additionally, Transformer\cite{Transformer} models are utilized to learn contextual information of point clouds, enabling the model to capture global feature information. However, DCP still exhibits certain limitations when dealing with point cloud data with significant initial pose differences. In the presence of added Gaussian noise\cite{added_noise}, although DCP\cite{dcp} demonstrates better robustness compared to methods like FGR\cite{FGR}, it is still subject to some degree of influence.

\textbf{An Adaptive Registration Method Based on Multimodal Data:  }The current trend in point cloud tasks is the increasing popularity of multimodal data\cite{pbsl}\cite{xu2021dual}. Geometry-based methods\cite{go_icp}\cite{RANSAC}\cite{SAC-IA}, in the context of point cloud registration, involve utilizing geometric features such as point positions, distances, and orientations to establish correspondences and achieve alignment between point clouds. These methods typically aim to find the optimal rigid transformation to minimize geometric disparities between point clouds, ensuring accurate registration. ICP (Iterative Closest Point)\cite{icp} and NDT (Normal Distributions Transform)\cite{ndt} are two of the most renowned geometry-based point cloud registration methods, particularly suitable for scenarios with local overlap and small-scale rigid transformations.

ICP (Iterative Closest Point)\cite{icp} is a classical method for point cloud registration, with the primary goal of iteratively finding the optimal rigid transformation between two point clouds to align them as closely as possible\cite{registration1}\cite{registration2}. The core idea of ICP involves iteratively mapping points from the target point cloud to the reference point cloud and updating the rigid transformation based on the corresponding mappings. This iterative process continues until convergence is achieved, ultimately realizing the best possible alignment between the two point clouds. NDT (Normal Distributions Transform)\cite{ndt} is a method used for point cloud registration, and its core idea involves describing the local structure of each point cloud by modeling the normal distribution of points. By mapping each point in the point cloud to its corresponding Gaussian distribution\cite{Gaussian_distribution}, NDT represents the point cloud as a set of probability density distributions\cite{probability_density_distribution}. During the registration process, the method adjusts the rigid transformation\cite{rigid_transformation} by minimizing the disparity in probability density distributions between the two point clouds, thereby achieving optimal point cloud alignment\cite{registration3}\cite{registration4}.

While both geometric-based methods\cite{icp}\cite{ndt}\cite{4pcs}\cite{SAC-IA} and their variants\cite{ef_icp}\cite{go_icp} have become increasingly mature, integrating them into practical engineering applications remains a highly challenging task. The ICP method performs well in scenarios with local overlap and small-scale rigid transformations but is sensitive to noise and prone to getting stuck in local optima. The NDT method demonstrates advantages in handling large-scale\cite{emrn}, sparse, or point clouds with complex geometric structures; however, its performance is constrained by the choice of parameters.

\section{Proposed Method}\label{sec3}
\subsection{Overview}
\begin{figure*}[t]
    \centering
      \includegraphics[width=1.0\linewidth]{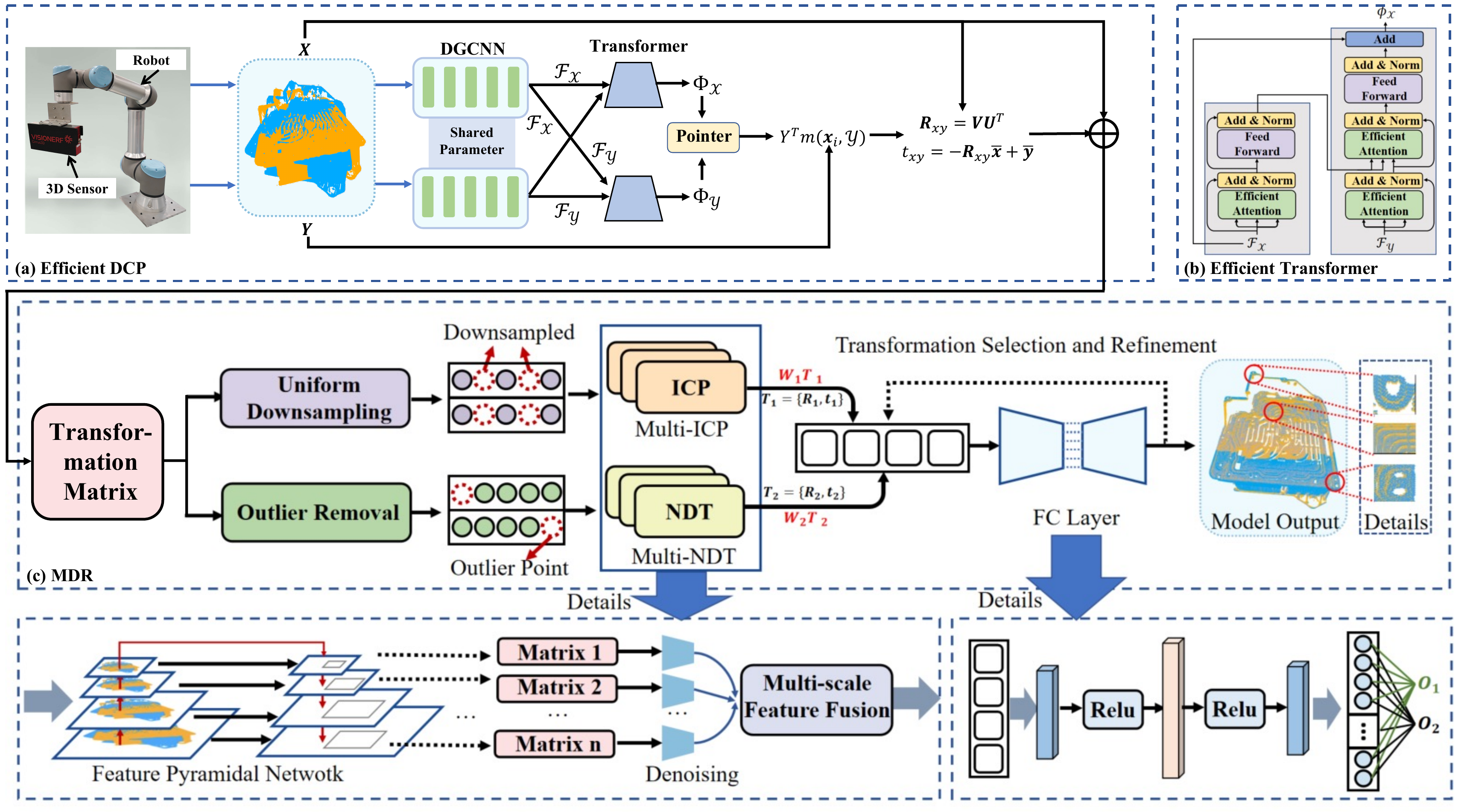}
    \caption{\small \textbf{The architecture of MEDPNet.}  In the diagram, part (a) shows the structure of Efficient DCP, part (b) illustrates the composition of Efficient Attention, and part (c) outlines the framework of the MDR method, with a detailed exposition of its details. The MEDPNet method collects point cloud data of the same die-casting part under different postures through a robotic arm equipped with a 3D sensor, inputs the unaligned point cloud pairs into Efficient DCP for preliminary registration, and then refines the alignment through MDR to preserve essential feature information.}
%\vspace{-.2cm}
    \label{fig:2}
 \end{figure*}

Given two sets of point clouds \( X = \{ x_i \in \mathbb{R}^3 \,|\, i = 1, \ldots, N \} \) and \( Y = \{ y_i \in \mathbb{R}^3 \,|\, i = 1, \ldots, M \} \), the objective is to estimate rigid transformations \( T_i = \{ R_i, t_i \} \) for \(i = 1, \ldots, N\) to align these two point clouds. 
In Fig \ref{fig:2}, we present the architecture of MEDPNet. In brief, we embed the acquired die-casting part point cloud data into a high-dimensional space using DGCNN\cite{DGCNN} and encode the contextual information with the Efficient Attention\cite{efficient} module, finally estimating the alignment using a differentiable SVD layer]\cite{svd}.
In which, \(x^P_i\) is the embedding of point \(i\) in the \(P\)-th layer, and \(h^P_\theta\) is a nonlinear function in the \(P\)-th layer parameterized by a shared multilayer perceptron (MLP). The forward mechanism is given by:
\begin{equation} 
x^P_i = h^P_\theta(x^{P-1}_i)
\label{eq:1} 
\end{equation} 

We input the collected unaligned input point clouds X and Y into the same space, where we embed each point of the two input point clouds individually, and iterate over the features of each point in the input point clouds, represented by the aggregation function as: 
\begin{equation}
F_x = \{x^P_1, x^P_2, \ldots, x^P_i, \ldots, x^P_N\},
\label{eq:2} 
\end{equation} 
and
\begin{equation}
F_y = \{y^P_1, y^P_2, \ldots, y^P_i, \ldots, y^P_N\}
\label{eq:3} 
\end{equation} 
DGCNN constructs a k-NN (k-nearest neighbor) graph \(M\), nonlinearly acquires edge values at edge endpoints, and performs per-vertex aggregation at each layer. Unlike PointNet\cite{pointnet}, which extracts independent information from each point, DGCNN explicitly incorporates local geometric shapes into its representation. This is achieved through the forward mechanism: 
\begin{equation}
x^P_i = f(\{h^P_\theta(x^{P-1}_i, x^{P-1}_j) \forall j \in N_i\})
\label{eq:3} 
\end{equation} 
where \(N_i\) represents the set of neighbors of point \(i\) in the k-NN graph, ensuring that local geometric features are considered during the aggregation process. In the task of die-casting part point cloud registration, DGCNN achieves higher quality registration performance by leveraging these local geometric information.

%图2

% 图：3
\begin{figure}[t]
    \centering
    \includegraphics[width=0.4\linewidth]{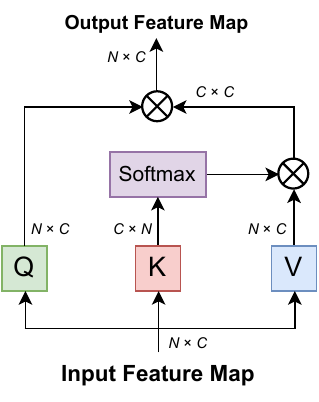}
\caption{\textbf{Illustration of the architecture of efficient attention}.Where the input feature map undergoes a transformation into three distinct components: Queries (\textit{Q}), Keys (\textit{K}), and Values (\textit{V}). These components facilitate a self-attention schema by computing attention scores between \textit{Q} and \textit{K}, followed by a softmax normalization to acquire a probabilistic weight distribution. The weighted sum of these probabilities with \textit{V} culminates in the output feature map, encapsulating a dynamic representation of salient features pivotal for subsequent layers of the network to process. This mechanism underpins the network's capacity to accentuate pertinent information within the feature space selectively.}
\label{fig:3}
\end{figure}

\subsection{Efficient Attention}
Before introducing Efficient attention, let's first discuss the concept of dot-product attention. Dot-product attention is a fundamental attention mechanism commonly used in models like Transformers. For a given query vector \( Q \), key vector \( K \), and value vector \( V \), the computation of dot-product attention is as follows:
\begin{equation}
\text{Attention}(Q, K, V) = \text{softmax}\left(\frac{QK^T}{\sqrt{d_k}}\right)V 
\label{eq:4} 
\end{equation} 
Here, \( d_k \) is the dimensionality of query/key vectors, \( QK^T \) represents the dot product between query vector \( Q \) and key vector \( K \), and it is scaled by \( \sqrt{d_k} \) to stabilize gradient magnitudes. The softmax function\cite{softmax} normalizes the dot product results into attention weights, which are then used to weight the value vector \( V \) to generate the final output. This mechanism allows the model to dynamically allocate attention based on the similarity between queries and keys, capturing relationships between different positions in the input sequence.

The principle of efficient attention is to optimize the traditional attention mechanism, particularly in addressing the challenges of high computational complexity and significant memory consumption when processing long sequence data. By reducing redundancies in computation and employing more efficient computational strategies, such as low-rank factorization and kernel techniques, it approximates the key operation \(QK^T\) in the standard self-attention mechanism, as shown in Fig \ref{fig:3}. 

In the standard self-attention mechanism, the input sequence \(X \in \mathbb{R}^{n \times d}\) undergoes linear transformations to obtain queries \(Q\), keys \(K\), and values \(V\), where \(Q = XW_Q\), \(K = XW_K\), \(V = XW_V\), and \(W_Q\), \(W_K\), \(W_V \in \mathbb{R}^{d \times d_k}\) are the corresponding weight matrices, with \(d_k\) being the feature dimension.

Traditionally, attention weights are obtained by computing \(QK^T\) and applying the softmax function, i.e., 
\begin{equation}
\text{Attention}(Q,K,V) = \text{softmax}\left(\frac{QK^T}{\sqrt{d_k}}\right)V
\label{eq:4} 
\end{equation} 
This step has a computational complexity of \(O(n^2d_k)\), which for long sequence data, results in significant computational burden and memory requirements.

Efficient attention introduces an approximation technique to reduce this complexity, specifically, it uses a form 
\begin{equation}
\text{Attention}(Q,K,V) \approx \text{softmax}\left(\frac{\phi(Q)\phi(K)^T}{\sqrt{d_k}}\right)V
\label{eq:4} 
\end{equation} 
for approximation, where \(\phi(\cdot)\) is a nonlinear function mapping to a lower-dimensional feature space, effectively reducing the required computational power and storage space. This mapping not only reduces the need for direct computation of \(QK^T\) but also, by selecting an appropriate \(\phi\) function, can lower the computational complexity of the attention mechanism from \(O(n^2d_k)\) to \(O(nmd_k)\), where \(m\) is the dimension of the mapped lower-dimensional space, significantly smaller than the length of the input sequence \(n\).
 
Precisely for these reasons, Efficient attention significantly enhances computational and storage efficiency in die-cast parts point cloud registration tasks by optimizing the attention mechanism. It effectively manages long-sequence dependencies, accurately captures changes and spatial relationships between point clouds, achieves high-precision registration, and broadens the application scope in resource-constrained environments.

\subsection{Adaptive Multi-scale Patch Matching for Registration}
First, let's review the geometric point cloud registration method. Its core idea involves iteratively optimizing the rigid transformation \{R, t\}, where \( R \) is the rotation matrix and \( t \) is the translation vector. The objective is to minimize the distance \( T \) between two point clouds, achieving their alignment. The expression for this can be given as follows:
\begin{equation}
T^* = \arg\min_{T} \sum\limits_{i=1}^{N} \lVert f(x_i) - y_i \rVert^2 
\label{eq:4} 
\end{equation} 
our approach primarily adopts the ICP (Iterative Closest Point) method and the NDT (Normal Distributions Transform) method. The core idea of the ICP method is to achieve point cloud registration by iteratively optimizing the rigid transformation \{R, t\} to minimize the distance \( T \) between point clouds. Its expression is:
\begin{equation}
T^* = \min_{R, t} \sum_i \left\| Rx_i + t - y_i \right\|^2 
\label{eq:4} 
\end{equation} 
Here, ICP primarily aligns point clouds through iterative optimization. In the process of minimizing the objective function, adjustments to \(R\) and \(t\) are made to bring the two point clouds as close together as possible in space.

% 图4
\begin{figure}[t]
    \centering
    \includegraphics[width=0.6\linewidth]{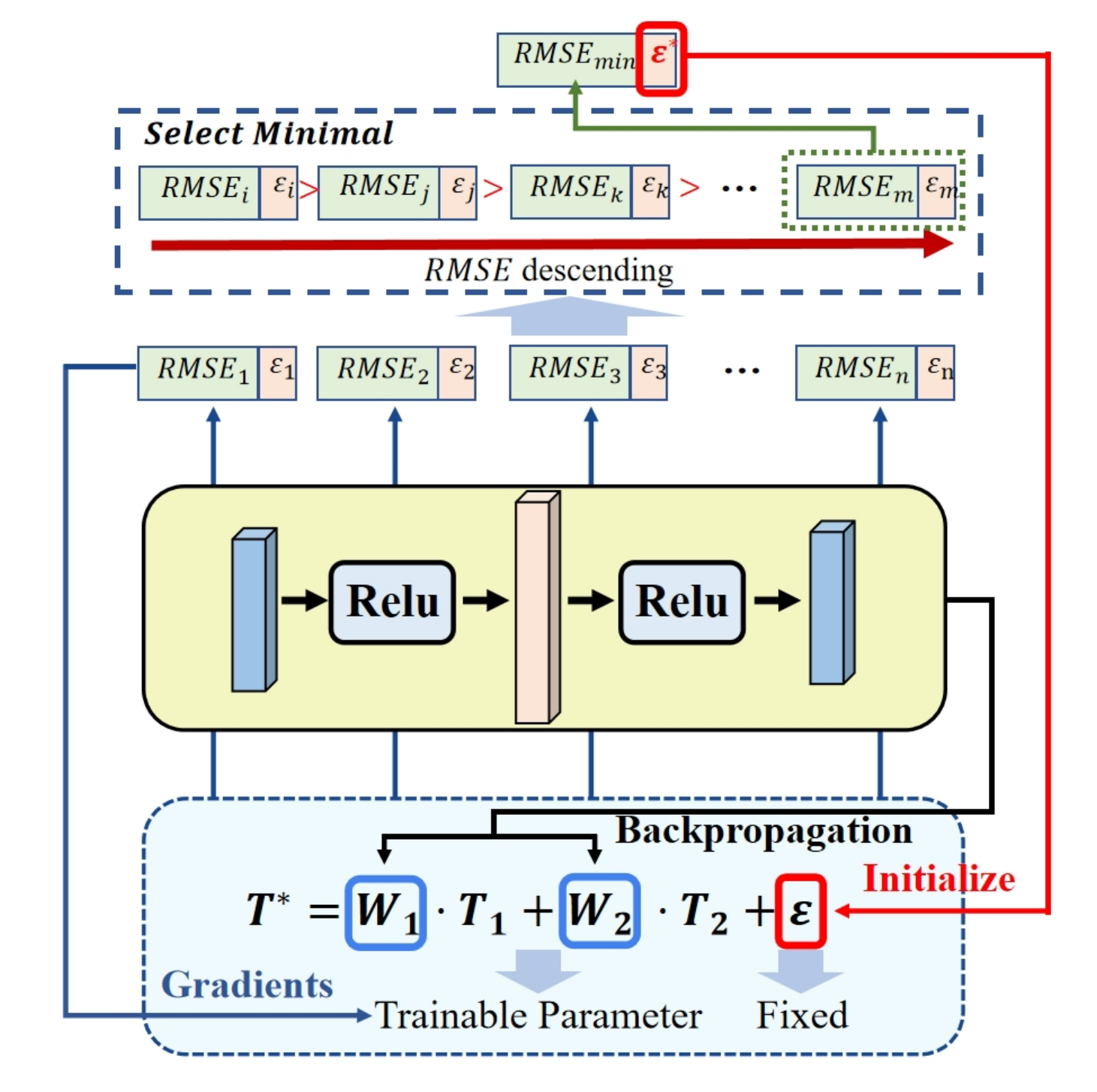}
\caption{\textbf{Adaptive parameter optimization}. First, we input the formula to be optimized into an MLP and use backpropagation to optimize the weights. Next, through a self-updating filtering mechanism, we iterate to find the smallest registration error, iteratively updating the adaptive hyperparameter \(\varepsilon\).}
%\vspace{-.2cm}
\label{fig:4}
\end{figure}

The NDT method describes the local structure of point clouds by modeling the normal distribution of each point. Its optimization objective is formulated as:
\begin{equation}
T^* = \min_{R, t} \sum_i \frac{1}{2} (\mu_{x_i} - \mu_{y_i})^{T}_{2} \Sigma_i^{-1} (\mu_{x_i} - \mu_{y_i}) 
\label{eq:4} 
\end{equation} 
\(\mu_{x_i}\) and \(\mu_{y_i}\) denote the mean of the normal distribution corresponding to points in two point clouds, and \(\Sigma_i\) represents the covariance matrix of the normal distribution. The goal is to adjust \(R\) and \(t\) to make the two point clouds as consistent as possible in terms of normal distribution, minimizing the objective function.

\textbf{Multi-scale Feature Fusion Module:}  In both ICP (Iterative Closest Point) and NDT (Normal Distributions Transform), we introduced a multiscale feature fusion module aimed at addressing key challenges in point cloud registration, such as local minima and sensitivity to initial values. Our approach first utilizes the Feature Pyramid Network (FPN) \cite{fpn} method  to generate multiple scales of point clouds through downsampling. At each scale, the ICP algorithm is independently applied to find the optimal rigid transformation. Subsequently, the coarse-scale registration results are propagated to finer scales, resulting in the final registration outcome and transformation matrix. We have successfully implemented feature fusion on die-cast point cloud pairs at \(k\) different scales. 

The multiscale ICP method aims to minimize the registration error across all scales \(k\) and corresponding points \(i\) by finding the optimal rigid transformation \(R_k\) and \(t_k\). This approach allows us to comprehensively tackle registration challenges at multiple scales, thereby enhancing the algorithm's robustness in diverse scale environments. The expression for this method is given by:
\begin{equation}
T^* = \min_{R_k, t_k} \sum_k \sum_i \|R_kx_i^k + t_k - y_i^k\|^2
\label{eq:4} 
\end{equation} 
similarly, multiscale Normal Distributions Transform (NDT) minimizes the cumulative error of the normal distribution for corresponding points at each scale, utilizing the Mahalanobis distance metric (measured by the difference between the inverse covariance matrix and the normal distributions). This optimization seeks to find the optimal rigid transformation \(R_k\) and translation vector \(T_k\) at each scale. This enables precise registration of point cloud \(X\) with target point cloud \(Y\) in the normal distribution across multiple scales through adjustments in rigid transformation and translation. It can be expressed as:
\begin{equation}
T^* = \min_{R_k, T_k} \sum_k \sum_i \frac{1}{2} (\mu_{x_i}^k - \mu_{y_i}^k)^T \Sigma_i^{-1} (\mu_{x_i}^k - \mu_{y_i}^k) 
\label{eq:4} 
\end{equation} 

\textbf{Dual Channel Fusion Module:}  Previous studies have primarily focused on enhancing the performance of the ICP (Iterative Closest Point) and NDT (Normal Distributions Transform) methods, yet they have overlooked the importance of ensuring stability under conditions of high precision. This issue becomes particularly evident when dealing with complex die-cast point cloud data, where the ICP algorithm may perform excellently on point cloud data "a," while the NDT algorithm shows superior performance on point cloud data "b." To overcome this challenge, we propose a novel dual-channel fusion module. Through multi-scale feature fusion, this module enables the ICP and NDT methods to obtain rigid transformation matrices \(T_1\) and \(T_2\), respectively. We input 300 pairs of rigid transformation matrices from the ICP and NDT methods into an MLP for learnable self-feedback weighting and iterate the weights of the optimal registration results by minimizing the registration error, as shown in Fig 4. The MLP consists of three fully connected layers, with neuron counts of 32, 64, and 32, respectively. Here, we opt for the Huber loss function, expressed as:
\begin{equation}
L_\delta(a) = 
\begin{cases} 
\frac{1}{2}a^2 & \text{for } |a| \le \delta,\\
\delta(|a| - \frac{1}{2}\delta) & \text{otherwise.}
\end{cases}
\label{eq:4} 
\end{equation}
Where \(a = l - \hat{l}\) represents the prediction error, i.e., the difference between the actual value \(l\) and the predicted value \(\hat{l}\). \(\delta\) is a threshold parameter that determines the point at which the loss function transitions from squared error to linear error. When the absolute value of the error is less than or equal to \(\delta\), the loss function behaves like the square of the error (similar to MSE), imposing a heavier penalty for smaller errors to encourage more precise fitting. Conversely, when the absolute value of the error exceeds \(\delta\), the loss function becomes linear (similar to MAE), reducing the penalty for larger errors and enhancing the model's robustness.

Although the merged matrix \(T\) obtained at this point demonstrates certain reliability, repeated tests have shown that the model's accuracy decreases when facing unfamiliar samples. We hypothesize that this instability might be due to the limited sample size, making it difficult for the model to learn the complete features of die-cast part point clouds. However, due to the irreplicability of die-cast samples and the industrial production cycle's inability to accommodate the training duration for a large volume of samples, we introduced the hyperparameter \(\varepsilon\). Initially, we hoped to directly obtain optimal weights and \(\varepsilon\) through the MLP mechanism, but the limited training samples and excessive number of parameters to be optimized led to unsatisfactory results. To address this, we added a self-updating filtering mechanism on top of the MLP. With the determination of optimal weights \(W_1^*\) and \(W_2^*\), we use the root mean square error (RMSE) feedback from each iteration to determine the corresponding \(\varepsilon\), choosing the \(\varepsilon\) associated with the minimum RMSE as the input for the next iteration, as shown in Fig \ref{fig:4}.The formula for RMSE is as follows:
\begin{equation}
RMSE = \sqrt{\frac{1}{N} \sum_{i=1}^{N} (l_i - \hat{l}_i)^2}
\label{eq:4} 
\end{equation} 
Where \(N\) is the number of samples, \(l_i\) is the actual value of sample \(i\), and \(\hat{l}_i\) is the predicted value for sample \(i\).Finally,we can obtain:
\begin{equation}
T^* = W_1^* \cdot T_1 + W_2^* \cdot T_2 + \mathbf{\varepsilon^*}
\label{eq:4} 
\end{equation} 
In this setup, we achieved learnable adaptive registration through a multilayer perceptron and a self-updating filtering mechanism, obtaining desirable results on the die-cast part point cloud dataset DieCastCloud.

\section{Experiment and Analysis}\label{sec:4}  
In this section, we will conduct comparative experiments to assess the effectiveness of our approach. We first introduce the details of the experiments in Section 4.1. In Section 4.2, we evaluate the Efficient DCP method on our die-cast dataset and perform ablation experiments to ensure the method's effectiveness. In Section 4.3, we introduce the an adaptive registration method based on multimodal data, providing corresponding experiments at each step. In Section 4.5, we will present an overall introduction to our method, MEDPNet (Multimodal Efficient Deep Closest Point), and compare it with the current state-of-the-art methods.

\subsection{Implementation Details}
This experiment utilizes Open3D\cite{Open3D} 1.2.0 and PCL 1.9.1 to implement algorithm execution in Python and C++. The experimental platform is Ubuntu 18.04 system, with PyTorch\cite{pytorch}
version 1.8.1, CUDA version 11.1, GPU=RTX 3090 (24GB) * 1, CPU=15 vCPU AMD EPYC 7642 48-Core Processor.
To test the generalization of different models, we will split DieCastCloud into training and testing sets.

Due to the complexity of the surface features of die cast parts, unlike the approach in PointNet\cite{pointnet} experiments of uniformly sampling 1024 points on the model's outer surface, we opted to sample 4096 points. This decision was based on an understanding of the complexity of die cast surface features, aiming to more comprehensively preserve the point cloud's feature information, thereby enhancing the accuracy and reliability of subsequent registration.

To ensure consistency and standardization in data processing, we performed a series of preprocessing steps on the collected point cloud data. Initially, the point cloud data was centered at the origin and scaled to fit within a unit sphere. Throughout this process, we only used the three-dimensional coordinates (x, y, z) of the points as input features, without introducing any additional attribute information, to accurately assess the model's ability to recognize and process geometric shapes themselves.

The initial pose has a critical impact on point cloud registration. To better quantify the performance of coarse registration, we employed multiple error metrics, including Mean Squared Error (MSE), Root Mean Squared Error (RMSE), and Mean Absolute Error (MAE), to ensure the reliability of the method. Moreover, considering practical applications in the die casting industry, we primarily focused on Root Mean Squared Error (RMSE) and registration time (s) during fine registration, where all angle measurements were made in degrees (°).

\subsection{Datasets}  
The DieCastCloud dataset contains 2,000 point cloud data, including 5 different types of die-cast parts. This dataset is randomly divided into a training set and a test set, with proportions of 0.8 and 0.2, respectively. In practical applications in the die-casting industry, the purpose of point cloud registration is to enhance the completeness of the point cloud data while preserving key features, to ensure high-precision 3D reconstruction and facilitate product quality control. Unlike other datasets\cite{ModelNet40}\cite{kitti}, here, to ensure the practical feasibility of the method, the overlap rate of point cloud data in DieCastCloud is set to be greater than \(85\%\).

We utilized the UR16e robotic arm equipped with the high-precision 3D laser scanner CIRRUS 3D 300 to collect point cloud data of die-cast parts, and we named the resulting dataset DieCastCloud. The point cloud data in DieCastCloud covers the main external surfaces of the die-cast parts, including complex geometric features such as pipes and holes. Additionally, we processed and filtered the collected raw point cloud data to obtain a richer sample set.

Finally, in the creation process of DieCastCloud, point cloud data was enhanced through techniques such as rotation, translation, scaling, and random erasure to increase the diversity of the data and enhance the model's generalization capabilities. Specifically, we randomly rotated the point cloud data at random angles along any axis and translated it in any spatial direction, with the translation range controlled within [-800mm, 800mm]. The scaling ratio was constrained to between [0.95, 1.05] of the original point cloud size.

\subsection{Efficient of DCP} 
In this experiment, we use DCP-v2, which incorporates a Transformer, as our baseline. We compare the performance of Efficient DCP with other cutting-edge deep learning-based point cloud registration methods, including PointNetLK\cite{pointnetlk}, GeoTransformer\cite{geo}, PRNet\cite{prnet}, DeepGMR\cite{deepgmr}, and DCP\cite{dcp}. We randomly split 2000 die-cast component point clouds from DieCastCloud into validation and test sets, utilizing different point cloud data during the training and testing periods.During training, we sampled the point clouds and applied a random rigid transformation along each axis, with rotations uniformly sampled within [0, 60°] and translations in the range of [-150mm, 150mm]. The source point cloud and the point cloud after the rigid transformation were used as the input to the network.

To ensure a fair comparison among these methods, we follow the convention and use performance metrics including Mean Squared Error (MSE) for rotation angles (MSE(R)), Mean Squared Error for translation directions (MSE(t)), Root Mean Squared Error for rotation angles (RMSE(R)), Root Mean Squared Error for translation directions (RMSE(t)), Mean Absolute Error for rotation angles (MAE(R)), and Mean Absolute Error for translation directions (MAE(t)), to guarantee the reliability of the experiments.

Table \ref{tab1} assesses the performance of our method and its counterparts in this experiment. In this study, Efficient DCP adopts a structure that integrates DGCNN with Efficient Transformer, with a learning rate of 0.001, 200 epochs, train batch sizes of 32 and train batch sizes of 10, and employs Stochastic Gradient Descent (SGD) as the optimizer. Across all evaluated performance metrics, Efficient DCP showcases outstanding performance.

\begin{table}[h]
\caption{Evaluating the performance of Efficient DCP against other advanced methods}\label{tab1}
\begin{tabular}{@{}lllllll@{}}
\toprule
Model     & MSE(R)      & MSE(t)      & RMSE(R)      & RMSE(t)      & MAE(R)      & MAE(t)  \\
\midrule
PointNetLK   & 51.271578     & 0.114432     & 6.842565     & 0.089526     & 7.664845     & 0.045454     \\
GeoTransformer & \underline{24.352470}    & \underline{0.001422}     & 5.898481     & \textbf{0.002177} & 2.974119     & 0.084997     \\
PRNet     & 82.665150     & 0.014432     & 12.54549     & 0.114551     & 4.859481     & 0.072361     \\
DeepGMR    & 29.159647     & 0.008747     & \underline{3.861095}     & 0.084411     & 3.784151     & 0.048944     \\ \hline
DCP      & 24.372444     & 0.009330     & 4.851226     & 0.017721     & \underline{2.311324}     & \underline{0.027983}     \\
Efficient DCP(Ours) & \textbf{4.822984} & \textbf{0.000231} & \textbf{2.196129} & \underline{0.015187}     & \textbf{1.350260} & \textbf{0.008338}\\
\botrule
\end{tabular}
\footnotetext{Performance metrics include Mean Squared Error (MSE), Root Mean Squared Error (RMSE), and Mean Absolute Error (MAE) in both translation and rotation directions. \textbf{Boldfaced} numbers highlight the best performance and the second best are \underline{underlined}.}
\end{table}

\subsection{Multiscale Feature Fusion Dual-channel Precision Registration} 
In this section, we conducted comparative experiments to validate our choice of the Iterative Closest Point (ICP) and Normal Distributions Transform (NDT) methods. In real industrial scenarios, registration time needs to be controlled within 60 seconds. To better evaluate the feasibility of the methods, we used Root Mean Square Error (RMSE) in millimeters and registration time in seconds as performance evaluation metrics, with the results shown in Table \ref{tab2}. Additionally, to more intuitively understand the impact of rotation angles (°) and translation distances (mm) on various methods, we conducted comparative experiments to test the performance of each method at specific rotation angles and translation distances, with results presented in Table \ref{tab3}. Finally, we elucidated the advantages of our dual-channel precision registration method based on multi-scale features.

\begin{table}[h]
\caption{Performance on clean and noisy samples}\label{tab2}
\begin{tabular*}{\textwidth}{@{\extracolsep\fill}lcccccc}
\toprule%
& \multicolumn{2}{@{}c@{}}{Data clean} & \multicolumn{2}{@{}c@{}}{Data noisy} \\\cmidrule{2-3}\cmidrule{4-5}%
Method & RMSE(mm) & Time(s) & RMSE(mm) & Time(s) \\
\midrule
SAC-IA          & \multicolumn{1}{c}{\underline{0.128}}    & 47.64   & \multicolumn{1}{c}{0.206}    & 53.76   \\
4PCS              & \multicolumn{1}{c}{0.168}    & 13.32   & \multicolumn{1}{c}{0.273}    & \underline{16.11}   \\
NDT               & \multicolumn{1}{c}{0.158}    & \textbf{4.33}    & \multicolumn{1}{c}{\underline{0.182}}    & \textbf{5.64}    \\
ICP               & \multicolumn{1}{c}{0.153}    & \underline{12.47}   & \multicolumn{1}{c}{0.197}    & 18.12   \\
MDR(ours)         & \multicolumn{1}{c}{\textbf{0.092}}    & 25.92   & \multicolumn{1}{c}{\textbf{0.148}}    & 29.41   \\
\botrule
\end{tabular*}
\footnotetext{Performance metrics primarily include Root Mean Square Error (RMSE) and registration time (s). \textbf{Boldfaced} numbers highlight the best performance and the second best are \underline{underlined}.}
\end{table}

\begin{table}[h]
\caption{Testing the impact of different rotation angles and translation distances on the experimental results}\label{tab3}
\begin{tabular*}{\textwidth}{@{\extracolsep\fill}lcccccc}
\toprule%
& \multicolumn{3}{@{}c@{}}{Data Rotate (°)} & \multicolumn{3}{@{}c@{}}{Data Translation (mm)} \\\cmidrule{2-4}\cmidrule{5-7}%
Method & 10 & 20 & 30 & 100 & 500 & 1000 \\
\midrule
SAC-IA                  & 0.004        & 0.008        & 0.015        & \underline{0.003}        & 0.007        & 0.015        \\
4PCS                    & \underline{0.002}        & 0.007        & 0.011        & 0.010        & 0.014        & 0.022        \\
NDT                     & 0.007        & 0.009        & 0.012        & \textbf{0.001} & \underline{0.003}        & \underline{0.010}        \\
ICP                     & \underline{0.002}        & \underline{0.004}        & \underline{0.008}        & \underline{0.003}        & 0.008        & 0.017        \\
MDR(ours)               & \textbf{0.001} & \textbf{0.001} & \textbf{0.003} & \textbf{0.001} & \textbf{0.002} & \textbf{0.002} \\
\botrule
\end{tabular*}
\footnotetext{We tested the Root Mean Square Error (RMSE) of each method when the rotation angles were 10°, 30°, and 90°, and the translation distances were 100mm, 500mm, and 1000mm, respectively. \textbf{Boldfaced} numbers highlight the best performance and the second best are \underline{underlined}.}
\end{table}

We first evaluated the performance of SAC-IA\cite{SAC-IA}, 4PCS\cite{4pcs}, NDT\cite{ndt}, ICP\cite{icp}, and MDR on the DieCastCloud dataset. To assess the robustness of these methods, we introduced noise with an intensity of 0.1 into the DieCastCloud dataset and then conducted comparative experiments to evaluate the stability of each method under noisy conditions. By incorporating a multi-scale feature fusion module, MDR possesses more comprehensive feature information compared to other methods, thereby achieving higher registration accuracy and noise resistance. As shown in Table \ref{tab2}, MDR demonstrates high performance while meeting the registration time requirements in practical applications.

%图5
 \begin{figure*}[t]
    \centering
      \includegraphics[width=1\linewidth]{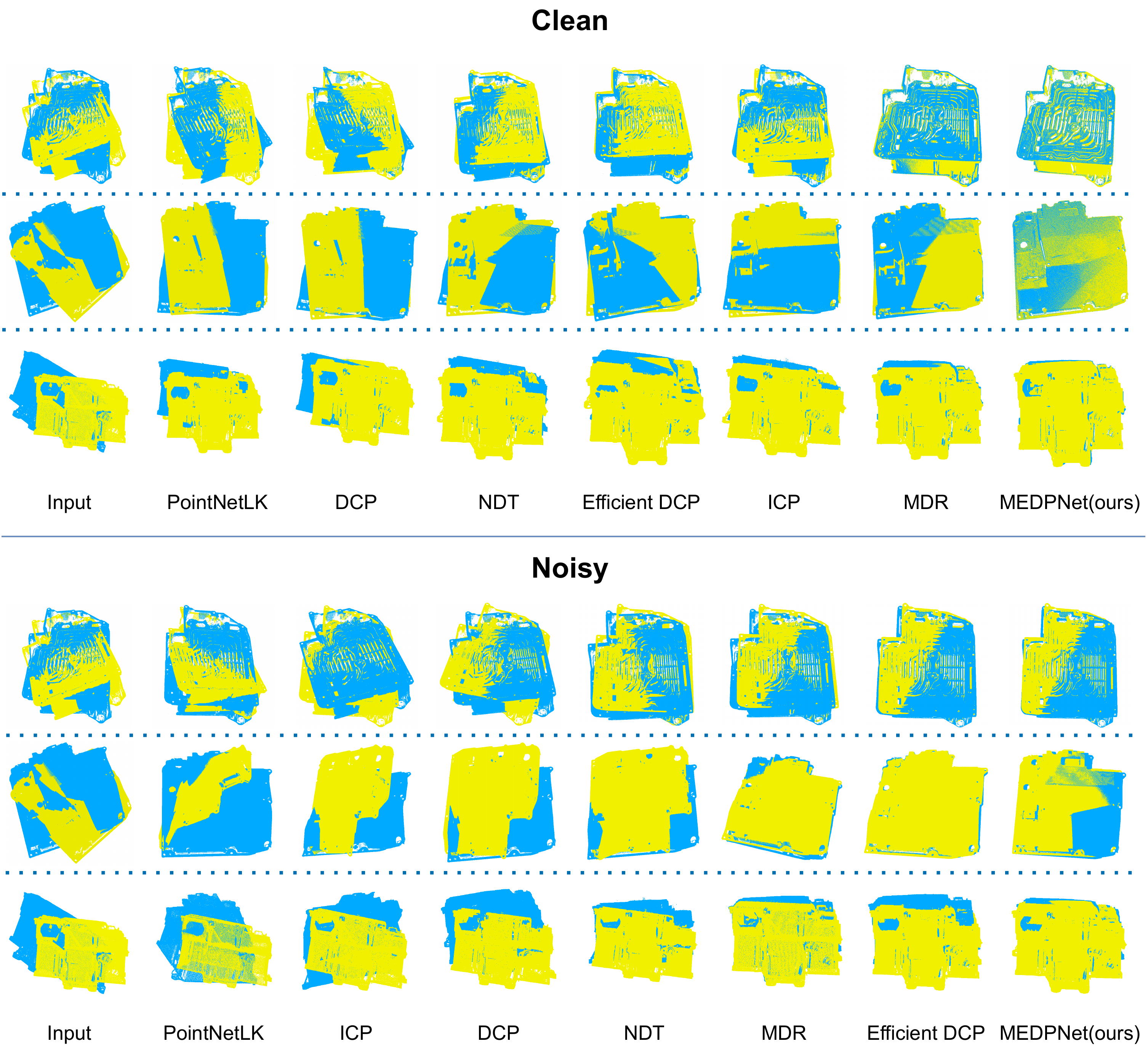}
    \caption{\small \textbf{Registration result visualization}. We visualized the registration results of various methods under clean and noisy samples, ranking them in descending order according to the root mean square error.}
%\vspace{-.2cm}
    \label{fig:5}
 \end{figure*}
 
Next, we tested the effects of rotation angles and translation distances on several algorithms. We rotated the point cloud data by 10°, 30°, and 90° around any spatial axis and translated it by 100mm, 500mm, and 1000mm in any spatial direction. As can be seen from Table \ref{tab3}, ICP showed a notable performance for different angles of rotation, while NDT performed better in facing translation issues and exhibited greater robustness in dealing with spatial position changes. Our method demonstrated the best registration performance compared to the other methods.

\begin{table}[h]
\caption{The registration performance of various methods under clean and noisy samples}\label{tab4}
\begin{tabular}{@{}lcc@{}} 
\toprule
Method & Data clean & Data noisy \\
\midrule
PointNetLK          & 6.84315 & 15.9744 \\
DCP                 & 4.85126 & 5.87391 \\
NDT                 & 4.27784 & 9.76239 \\
Efficient DCP (Ours) & 2.19618 & \underline{3.71646} \\
ICP                 & 2.07729 & 6.73248 \\
MDR (Ours)          & \underline{1.94877} & 4.67442 \\
MEDPNet (Ours)      & \textbf{1.17245} & \textbf{1.32954} \\ 
\botrule
\end{tabular}
\footnotetext{Here, we arrange the methods in descending order according to their root mean square error (RMSE) on clean samples. \textbf{Boldfaced} numbers highlight the best performance and the second best are \underline{underlined}.}
\end{table}

\subsection{Influence of MEDPNet}
Overall, the MEDPNet method achieves high-quality registration results by initially applying Efficient DCP for coarse registration of unaligned point cloud pairs, followed by fine-tuning through MDR. To assess the accuracy and robustness of our method, we conducted tests on both clean and noisy samples, with our experimental results presented in Table \ref{tab4}. We selected root mean square error (RMSE) as the performance metric to comprehensively account for both variance and bias. The experimental outcomes indicate that our method not only ensures high accuracy but also maintains robustness across different conditions.

\subsection{Visualization}  
In this section, we present a visual comparison of the performance of various advanced methods against our MEDPNet approach, as illustrated in Fig \ref{fig:5}. We selected three typical types of die casting samples and conducted experiments under both noise-free and noisy conditions to ensure the practicality of our method. Our comparison mainly includes PointNetLK\cite{pointnetlk}, DCP\cite{dcp}, ICP\cite{icp}, NDT\cite{ndt}, and our improved methods Efficient DCP, MDR, and MEDPNet, arranged in descending order according to root mean square error (RMSE). As can be observed in Fig \ref{fig:5}, MEDPNet achieves state-of-the-art performance on both clean and noisy samples.

\section{Conclusion}\label{sec5} 
In the intricate domain of die casting, where complex spatial structures and heterogeneous geometric features prevail, the quest for precise and resilient point cloud registration represents a formidable challenge. Traditional methodologies predominantly hinge on high-caliber datasets, endeavoring to enhance registration fidelity through network model optimization, yet frequently neglecting the nuances of real-world deployment.
Addressing this lacuna, the present exposition delineates the Multiscale Efficient Deep Closest Point (MEDPNet) modality, coupled with the establishment of DieCastCloud, a bespoke point cloud dataset specifically designed to mitigate the application impediments of point cloud registration within the die casting sphere.

The MEDPNet initially conducts coarse registration based on Efficient-DCP, subsequently transitioning to advanced precision registration via the Multiscale feature fusion dual-channel registration (MDR) method. By replacing the traditional Transformer's attention mechanism with Efficient Attention, it introduces a Multiscale feature fusion dual-channel precision registration (MDR) technique. This technique minimizes registration errors by adaptively optimizing the final transformation matrix using multilayer perceptrons (MLP), resulting in an adaptive, scalable, and highly robust global point cloud registration framework.

Although our method achieves excellent registration results for die casting point clouds, there are still some shortcomings. Firstly, due to the large size of Die Castings, as well as the presence of occlusions, blind spots, and the influence of machining marks, collecting high-quality point cloud data often requires a significant amount of time. Furthermore, for different Die Castings, it is necessary to adjust the data collection strategy of the robot. We believe that an intelligent die casting point cloud generator is key to solving this problem. Secondly, in the step of precise registration, we utilize unsubsampling point cloud data, where the parameter count of each point cloud reaches the tens of millions level. In actual industrial production, how to reduce computational costs is an urgent problem that needs to be addressed.

\section{Acknowledgements}\label{sec6} 
This work was supported by both the Unveiling the Top Technical Research Project of Dalian City (2023JB11GX001) and the Key Special Projects of the National Key R\&D Program (2022YFB3706802).


\end{document}